\documentclass[10pt,twocolumn,letterpaper]{article}
\usepackage[pagenumbers]{cvpr}
\usepackage{multirow,pifont}
\newcommand{\cmark}{\ding{51}}
\newcommand{\xmark}{\ding{55}}
\DeclareMathOperator*{\argmin}{arg\,min}
\definecolor{cvprblue}{rgb}{0.21,0.49,0.74}
\usepackage[pagebackref,breaklinks,colorlinks,allcolors=cvprblue]{hyperref}
\usepackage[capitalize]{cleveref}
\def\confName{CVPR}\def\confYear{2027}
\title{Rethinking Post-Training Quantization for Large Vision-Language Models}
\author{
Minchan Kang\textsuperscript{1}, Kyeonghye Park\textsuperscript{1}, Seoyoung Cho\textsuperscript{1}, Junhee Lee\textsuperscript{1}, Daeshik Kim\textsuperscript{1}\thanks{Equal correspondence: Daeshik Kim and Yucheol Cho.}, Yucheol Cho\textsuperscript{2}\footnotemark[1] \\
\textsuperscript{1}Korea Advanced Institute of Science and Technology (KAIST), \textsuperscript{2}Hanbat National University \\
{\small\texttt{\{mc.kang,pkhpjhs,52tjdud,bedro1210,daeshik\}@kaist.ac.kr; yccho@hanbat.ac.kr}}
}

\newsavebox{\tablefitbox}
\newcommand{\fitTable}[2]{%
  \sbox{\tablefitbox}{#2}%
  \ifdim\wd\tablefitbox>#1\relax
    \resizebox{#1}{!}{\usebox{\tablefitbox}}%
  \else
    \usebox{\tablefitbox}%
  \fi
}

\begin{document}
\maketitle
\begin{abstract}
Post-training quantization (PTQ) compresses large vision-language models (LVLMs) by calibrating quantization parameters on a small dataset without retraining. However, despite increasingly sophisticated calibration strategies, prior methods suffer substantial performance degradation under low-bit weight--activation quantization. To understand this degradation, we locate where quantization error is concentrated and examine the corresponding weight and activation distributions. We observe that a few outliers dominate the ranges shared by per-channel (PC) weights and per-token (PT) activations, leaving insufficient resolution for most values and mapping many of them to zero. These findings motivate a shift in focus from calibration refinement to the underlying quantization representation. We therefore present Rotate-and-Refine Quantization (RRQ) as a simple, effective baseline for LVLM PTQ, combining local Hadamard transforms with group quantization and efficient group-aware execution. Across 8 multimodal tasks and diverse quantization settings, RRQ substantially outperforms baselines while reducing inference time and memory usage. At W4A4, it retains over 96\% of full-precision performance on 8B models and over 99\% on a 32B model, compared with less than 8\% for these baselines.
The source code is publicly available at \url{https://github.com/kmc3661/RRQ}.
\end{abstract}

\section{Introduction}
Large language models (LLMs) have demonstrated strong language understanding and reasoning capabilities across diverse tasks~\cite{brown2020,chowdhery2023,touvron2023}. Building on these capabilities, large vision-language models (LVLMs) connect visual encoders to language models, enabling strong performance in visual question answering, captioning, and multimodal reasoning~\cite{blip2,liu2023visual,dai2023instructblip}. However, these advances come with substantial computation and memory costs, motivating research on knowledge distillation~\cite{hinton2015distilling,li2023distilling}, token pruning and merging~\cite{rao2021dynamicvit,bolya2022token,kang2026text}, and quantization~\cite{jacob2018quantization,zhou2016dorefa} to reduce these costs while preserving performance.

\begin{figure}[t]
\centering\includegraphics[width=\columnwidth]{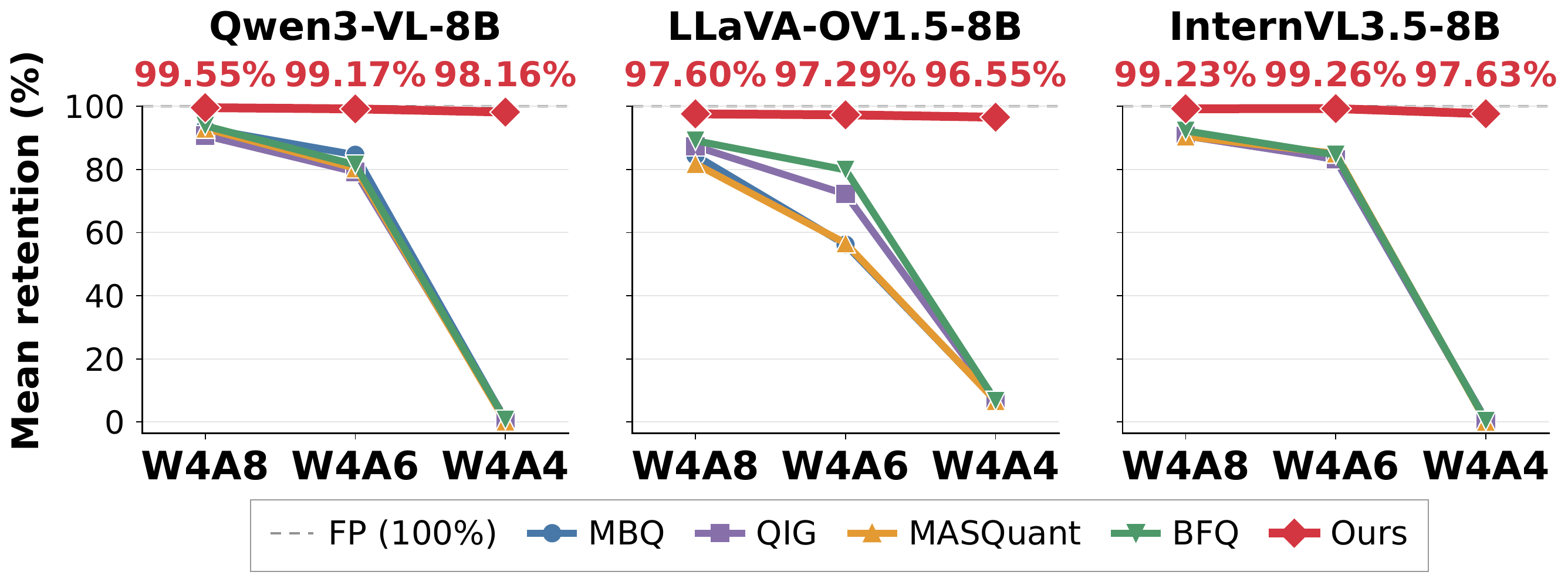}
\caption{Performance retention relative to full precision across 8 multimodal tasks on three 8B LVLMs~\cite{bai2025qwen3,an2025llava,wang2025internvl3}, compared with state-of-the-art LVLM PTQ methods~\cite{mbq,qig,hu2026masquant,kang2026beyond}.}
\label{fig:retention}
\end{figure}
\begin{figure*}[t]
\centering\includegraphics[width=\textwidth]{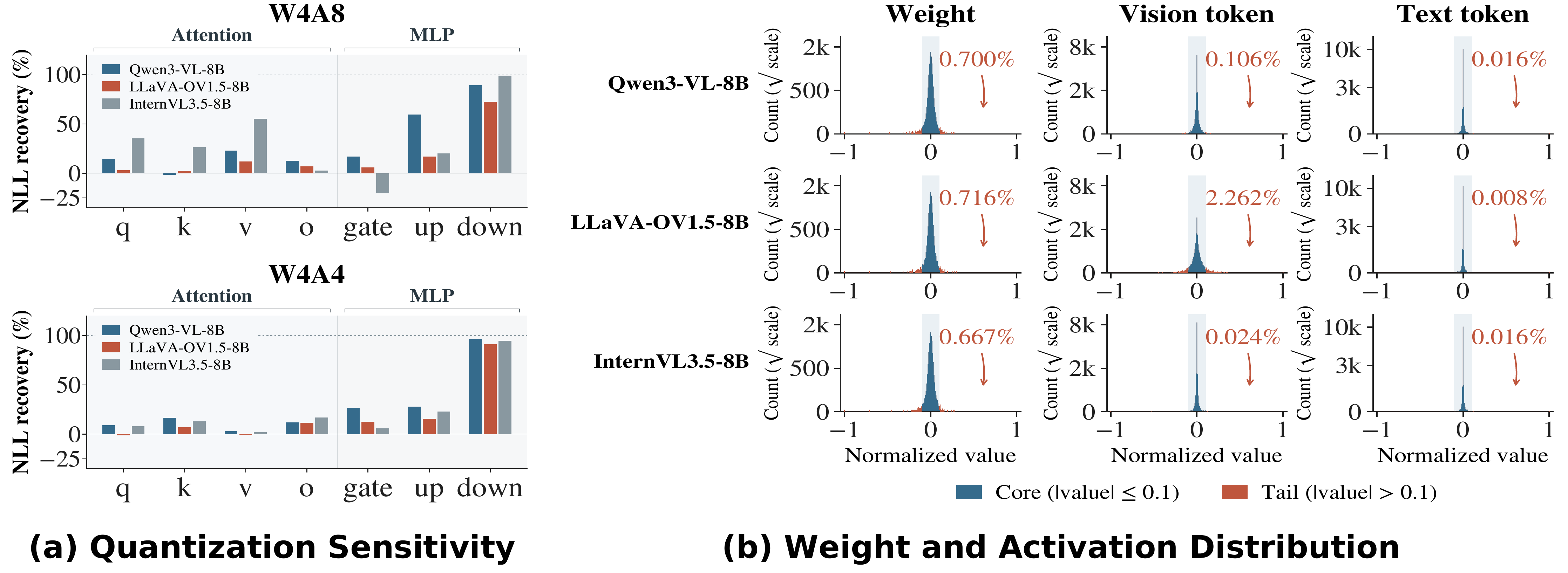}
\caption{\textbf{Outlier distributions in quantization-sensitive modules across three LVLMs}~\cite{bai2025qwen3,an2025llava,wang2025internvl3}. (a) Recovery of the quantization-induced increase in negative log-likelihood (NLL) on 32 COCO image--caption pairs~\cite{lin2014microsoft} when restoring one projection type at a time to FP weights and activations in a PC/PT-quantized baseline model. (b) Selected within-channel weight and within-token vision/text activation distributions in down projections, which yield the largest joint recovery in (a). Orange labels show the percentage of outliers whose magnitudes exceed 10\% of each vector's maximum absolute value.}
\label{fig:motivation1}
\end{figure*}

Among these approaches, post-training quantization (PTQ) offers a practical way to reduce computational and memory costs by quantizing weights and activations using only a small calibration set~\cite{nagel2020up,xiao2023smoothquant}. Following its success in LLMs~\cite{frantar2022gptq,lin2024awq,xiao2023smoothquant}, recent work has extended PTQ to LVLMs by accounting for the different characteristics of visual and textual representations. For example, MBQ balances modality-wise reconstruction using gradient-based sensitivity~\cite{mbq}, while QIG refines this treatment through token-level integrated gradients~\cite{qig}. MASQuant combines modality-aware smoothing with cross-modal compensation~\cite{hu2026masquant}, and BFQ adapts fitting strength to layer- and component-wise quantization effects~\cite{kang2026beyond}. Despite their increasingly fine-grained calibration strategies, performance retention still drops sharply as activation precision decreases, as shown in \cref{fig:retention}.

To understand this degradation, we investigate where quantization error is concentrated and find that certain module types are particularly affected. Examining their weight and activation distributions reveals a few outliers with substantially larger magnitudes than other weights within the same channel or other activations within the same token, as illustrated in \cref{fig:motivation1}. These outliers are particularly problematic for existing methods~\cite{mbq,qig,hu2026masquant,kang2026beyond}, which use per-channel (PC) weight and per-token (PT) activation quantization. Because the weights within each channel and the activations within each token share a single quantization range, the outliers widen this range, leaving insufficient resolution for other values and mapping many of them to zero.

Taken together, these observations point to the granularity of weight--activation quantization as a major bottleneck at low precision. Although modality-aware calibration can prioritize sensitive components, it remains constrained by the resolution available under PC/PT. Consequently, when many values already share the same quantization code, refining their relative importance alone cannot reliably preserve their distinctions. This motivates improving the underlying representation before further refining calibration strategies.

We therefore present Rotate-and-Refine Quantization (RRQ), a simple, strong baseline that combines local Hadamard transforms~\cite{ashkboos2024quarot} with group quantization~\cite{shen2020q,zhao2024atom}. The Hadamard transform redistributes values within small feature blocks to reduce outlier-dominated ranges. Group quantization then accommodates residual range differences with group-specific scales, preserving finer resolution in groups without outliers. To make this finer representation efficient, we combine Hadamard transformation and activation quantization in a single kernel to reduce memory traffic, and integrate group-wise corrections into integer accumulation to reduce computation overhead.

With these improvements in representation and execution, our baseline uses only a basic reconstruction-based scale search for calibration. It substantially outperforms prior LVLM PTQ methods across diverse quantization settings on 8 multimodal tasks. At W4A4, it retains over 96\% of full-precision performance on three 8B models and 99.06\% on a 32B model, with time and memory savings comparable to PC/PT. Our contributions are:
\begin{itemize}
\item We identify outlier-dominated PC/PT ranges as a key bottleneck behind low-bit performance collapse in LVLMs.
\item We establish a strong baseline combining Hadamard transforms, group quantization, and efficient execution with simple calibration.
\item We demonstrate strong performance retention and efficiency across diverse LVLMs and multimodal tasks.
\end{itemize}

\begin{figure*}[t]
\centering\includegraphics[width=\textwidth]{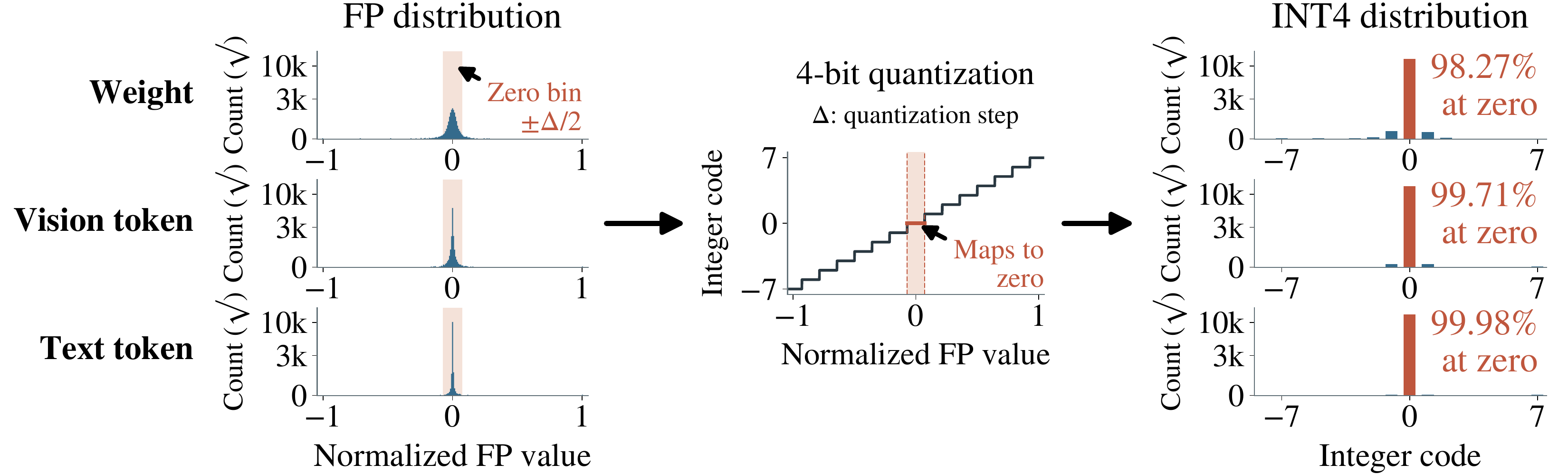}
\caption{\textbf{Zero-bin collapse under per-channel weight and per-token activation quantization.} The examples show 4-bit quantization of Qwen3-VL~\cite{bai2025qwen3} weight and activation distributions in \cref{fig:motivation1}(b). Most values within each weight channel or vision/text token map to zero.}
\label{fig:motivation2}
\end{figure*}

\section{Related work}
\subsection{Vision-language models}
LVLMs integrate LLMs~\cite{opt,le2022bloom} with pretrained vision encoders~\cite{dosovitskiy2020image,clip} through learned connectors. Models such as Flamingo~\cite{flamingo}, LLaVA~\cite{liu2023visual}, and InstructBLIP~\cite{dai2023instructblip}, along with subsequent LVLM families~\cite{qwen2,llavaov,internvl}, support multimodal understanding and instruction following. These advances rely on large language backbones and additional visual tokens, increasing memory and computational costs.

\subsection{Post-training quantization for LVLMs}
LLM PTQ includes weight-only and weight--activation quantization. Weight-only quantization methods use error compensation~\cite{frantar2022gptq} or activation-aware scaling~\cite{lin2024awq}. Weight--activation quantization reduces error through mixed precision~\cite{dettmers2022gpt3}, fine-grained quantization~\cite{yao2022zeroquant}, activation smoothing~\cite{xiao2023smoothquant}, learnable scaling and clipping~\cite{shao2024omniquant}, Hadamard rotations~\cite{ashkboos2024quarot}, and rotations with permutations~\cite{lin2024duquant}. Efficient kernels support group quantization~\cite{zhao2024atom} and progressive quantization~\cite{lin2025qserve}.

For LVLMs, Q-VLM~\cite{wang2024q} models cross-layer dependencies to guide quantization. MBQ~\cite{mbq} balances modality sensitivity, QIG~\cite{qig} estimates token sensitivity using integrated gradients, and MASQuant~\cite{hu2026masquant} combines modality-aware smoothing with cross-modal compensation. BFQ~\cite{kang2026beyond} balances fitting effort using component sensitivity. In contrast, we mitigate low-bit quantization loss with Hadamard transforms, group quantization, and uniform channel-wise equalization (CWE) scale search, without modality-specific calibration.

\section{Method}
\subsection{Preliminaries}
\label{sec:preliminaries}
Channel-wise equalization (CWE) is widely used~\cite{mbq,qig,hu2026masquant,kang2026beyond} to reduce quantization error by rescaling weights and activations. For a linear layer with activations $X$ and weights $W$, it applies a diagonal scale $D$ as $X_D=XD^{-1}$ and $W_D=WD$. This preserves $XW^{\mathsf T}$ while changing quantization error. We form candidate set $\mathcal D$ from normalized powers of channel-wise mean absolute activations, with uniformly spaced exponents in $[0,1]$, and select
\begin{equation}
D^*=\argmin_{D\in\mathcal D}\operatorname{MAE}\!\left(Q_A(X_D)Q_W(W_D)^{\mathsf T},XW^{\mathsf T}\right).
\label{eq:cwe_objective}
\end{equation}
Here $Q_A$ and $Q_W$ return dequantized activation and weight values, and MAE denotes mean absolute error.

\subsection{Motivation}
\label{sec:motivation}
Existing post-training quantization (PTQ) methods for large vision-language models (LVLMs)~\cite{mbq,qig,hu2026masquant,kang2026beyond} employ complex modality-aware calibration strategies, yet suffer substantial performance degradation under low-bit per-channel (PC) weight and per-token (PT) activation quantization.

To investigate this limitation, we use uniform CWE search as a basic calibration baseline for PC/PT quantization on 3 recent LVLMs~\cite{bai2025qwen3,an2025llava,wang2025internvl3} with different architectures. We restore the weights and input activations of each projection type to full precision across all layers, while keeping the other projections quantized. On 32 randomly sampled COCO image--caption pairs~\cite{lin2014microsoft}, we measure how much this restoration reduces the increase in reference-response negative log-likelihood (NLL) relative to the full-precision model. As shown in \cref{fig:motivation1}(a), restoring only the down projections recovers over 90\% of this NLL increase in all three models at W4A4, identifying these modules as a major source of degradation in this setting.

\Cref{fig:motivation1}(b) examines selected distributions in these modules: weights within the same output channel and activations within the same vision or text token. Although 97.7--99.99\% of values lie within 10\% of each vector's maximum absolute value, a few outliers extend the range far beyond this dense core. Under PC/PT quantization, these outliers determine the range shared by all values in the channel or token. For a vector $v$ with entries $v_i$, symmetric $b$-bit quantization uses a nominal step $\Delta(v)=\max_i|v_i|/(2^{b-1}-1)$.
An outlier-driven increase in $\Delta(v)$ therefore sacrifices resolution for the small values, with those satisfying $|v_i|<\Delta(v)/2$ rounding to zero. \Cref{fig:motivation2} shows this effect when 4-bit PC/PT quantization is applied to the Qwen3-VL examples in \cref{fig:motivation1}(b): 98.27\% of weights and 99.71\%/99.98\% of vision/text activations map to zero.

These observations identify outlier-driven loss of quantization resolution within weight channels and activation tokens as a major contributor to low-bit LVLM degradation. To address this limitation, we combine Hadamard transforms with group quantization in an integrated pipeline that preserves finer resolution with efficient execution.

\begin{figure*}[t]
\centering\includegraphics[width=\textwidth]{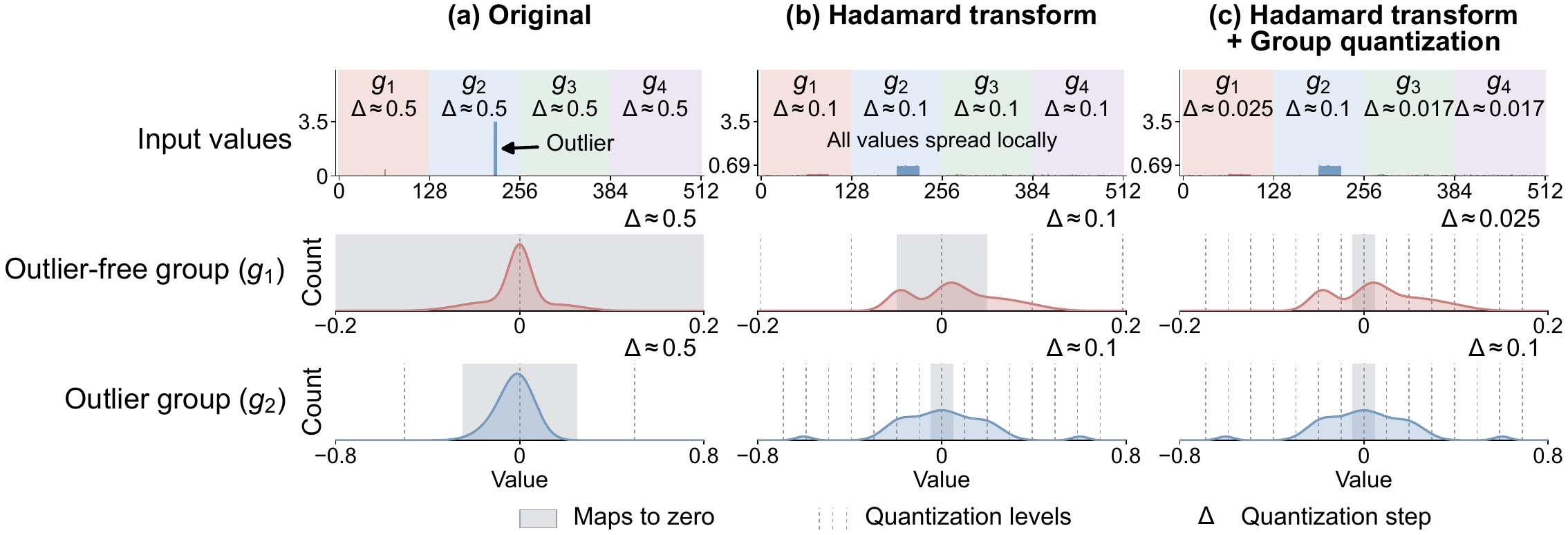}
\caption{\textbf{Local Hadamard transform and group quantization.} Based on an activation sample from \cref{fig:motivation1}, the illustration shows how local Hadamard transforms and group quantization affect an outlier group ($g_2$) and outlier-free groups ($g_1,g_3,g_4$). (a) Outliers cause coarse per-channel/per-token quantization. (b) Hadamard redistributes values locally, but a shared step remains coarse for outlier-free groups. (c) Group-specific scales refine their resolution. The middle and bottom rows illustrate $g_1$ and $g_2$, respectively.}
\label{fig:method}
\end{figure*}

\subsection{Rotate-and-Refine Quantization}
\paragraph{Hadamard transform.}
To reduce outlier-dominated ranges, we apply local Hadamard transforms~\cite{ashkboos2024quarot}. \Cref{fig:method} illustrates their effects on an outlier group ($g_2$) and outlier-free groups ($g_1,g_3,g_4$). Within each block of $h$ input features, $H_h$ applies an orthogonal rotation:
\begin{equation}
\widetilde{x}_j=\frac{1}{\sqrt{h}}\sum_{k=1}^{h}s_{kj}x_k,\qquad s_{kj}\in\{-1,+1\}.
\label{eq:hadamard_components}
\end{equation}
Here, $s_{kj}$ determines the signs. Each input contributes $\pm x_k/\sqrt{h}$ to every output, spreading its influence across the block.

We apply $R=\operatorname{diag}(H_h,\ldots,H_h)$ to both rescaled operands and compute quantization scales from $X_DR$ and $W_DR$. Orthogonality preserves their full-precision product, $(X_DR)(W_DR)^{\mathsf T}=XW^{\mathsf T}$, without an inverse transform.

\paragraph{Group quantization.}
As shown in \cref{fig:method}(b), the outlier group can still dominate the shared PC/PT range after Hadamard transformation, leaving an overly coarse step for outlier-free groups. We therefore divide each transformed weight row and activation token into groups of $G$ features and assign group-specific scales~\cite{dai2021vs,zhao2024atom}, adapting resolution to each local distribution as shown in \cref{fig:method}(c). Activation values in each group are quantized as
\begin{equation}
\begin{aligned}
\Delta_g&=\max_{k\in g}|z_k|/Q_a,\qquad \hat z_k=\Delta_gq_k,\\
q_k&=\operatorname{clip}(\operatorname{round}(z_k/\Delta_g),-2^{b_a-1},Q_a).
\end{aligned}
\label{eq:group_activation}
\end{equation}
where $z_k$ is an activation in group $g$, $b_a$ is its bitwidth, and $Q_a=2^{b_a-1}-1$. The step $\Delta_g$ maps values to codes $q_k$ and reconstructions $\hat z_k$; $\operatorname{clip}$ clamps to the specified bounds. Weights use group-affine quantization:
\begin{equation}
\begin{aligned}
u_k&=\operatorname{clip}(\operatorname{round}(w_k/s_g)+z_g,0,2^{b_w}-1),\\
\hat w_k&=s_g(u_k-z_g).
\end{aligned}
\label{eq:group_weight}
\end{equation}
where $w_k$ is a transformed weight, $b_w$ its bitwidth, $s_g$ the group scale, and $z_g$ the integer zero point. Here $u_k,\hat w_k$ denote codes and reconstructions.

With Hadamard transformation and group quantization incorporated into calibration, we use a uniform CWE search to select the scale that minimizes output reconstruction error for each quantized layer.

\subsection{Efficient group-aware execution}
\label{sec:efficient_execution}
Although RRQ mitigates outlier-induced quantization error, its online Hadamard transformation and group-specific scale handling introduce runtime overhead. For activation preparation, we fuse Hadamard transformation, group quantization, and packing into one kernel, producing packed activation codes and group-scale metadata for the subsequent linear operation. This avoids writing and rereading separate transformed-activation and unpacked-code buffers.

\begin{table*}[t]\centering
\caption{Results on 3 8B LVLMs~\cite{bai2025qwen3,an2025llava,wang2025internvl3} across 8 tasks. Ret.: mean FP retention. Bold: best quantized results.}\label{tab:main8}
\fitTable{\textwidth}{%
\begin{tabular}{cllrrrrrrrrr}\toprule
Model & Bits & Method & AI2D & ChartQA & MMStar & MMMU & TextVQA & VizWiz & NoCaps & TextCaps & Ret. (\%)\\\midrule
\multirow{16}{*}{\textbf{Qwen3-VL-8B}} & FP16 & -- & 83.94 & 85.32 & 67.20 & 60.00 & 83.78 & 69.33 & 65.09 & 82.15 & 100.00\\\cmidrule(lr){2-12}
 & \multirow{5}{*}{W4A8} & MBQ~\cite{mbq} & 76.59 & 78.80 & 62.00 & 52.67 & 77.68 & 67.93 & 61.23 & 78.69 & 93.02\\
 &  & QIG~\cite{qig} & 76.78 & 77.24 & 59.60 & 48.67 & 76.06 & 66.99 & 60.21 & 77.26 & 90.72\\
 &  & MASQuant~\cite{hu2026masquant} & 79.31 & 80.32 & 63.67 & 50.11 & 77.77 & 64.56 & 64.31 & 75.58 & 92.95\\
 &  & BFQ~\cite{kang2026beyond} & 78.30 & 79.08 & 63.80 & 53.56 & 77.93 & 68.90 & 59.56 & 79.25 & 93.82\\
 &  & \textbf{Ours} & \textbf{82.80} & \textbf{84.32} & \textbf{67.13} & \textbf{58.67} & \textbf{82.90} & \textbf{70.42} & \textbf{65.23} & \textbf{82.60} & \textbf{99.55}\\
\cmidrule(lr){2-12}
 & \multirow{5}{*}{W4A6} & MBQ~\cite{mbq} & 68.81 & 66.96 & 54.13 & 44.67 & 70.81 & 62.62 & 59.86 & 77.86 & 84.63\\
 &  & QIG~\cite{qig} & 64.28 & 62.24 & 50.87 & 40.33 & 68.57 & 61.44 & 51.61 & 75.11 & 79.21\\
 &  & MASQuant~\cite{hu2026masquant} & 74.51 & 57.60 & 55.27 & 45.67 & 59.24 & 58.33 & 58.00 & 69.33 & 80.37\\
 &  & BFQ~\cite{kang2026beyond} & 69.30 & 63.96 & 52.13 & 42.00 & 71.87 & 64.55 & 53.42 & 70.41 & 81.47\\
 &  & \textbf{Ours} & \textbf{82.06} & \textbf{84.76} & \textbf{67.13} & \textbf{58.56} & \textbf{82.67} & \textbf{69.25} & \textbf{65.16} & \textbf{82.25} & \textbf{99.17}\\
\cmidrule(lr){2-12}
 & \multirow{5}{*}{W4A4} & MBQ~\cite{mbq} & 0.58 & 0.00 & 0.47 & 2.00 & 0.00 & 0.00 & 0.02 & 0.14 & 0.61\\
 &  & QIG~\cite{qig} & 2.14 & 0.00 & 1.33 & 2.11 & 0.00 & 0.00 & 0.02 & 0.14 & 1.03\\
 &  & MASQuant~\cite{hu2026masquant} & 0.36 & 0.00 & 0.13 & 0.56 & 0.00 & 0.00 & 0.00 & 0.00 & 0.20\\
 &  & BFQ~\cite{kang2026beyond} & 1.85 & 0.00 & 0.93 & 1.11 & 0.00 & 0.00 & 0.02 & 0.10 & 0.70\\
 &  & \textbf{Ours} & \textbf{81.19} & \textbf{82.68} & \textbf{65.53} & \textbf{56.67} & \textbf{82.06} & \textbf{68.54} & \textbf{65.50} & \textbf{84.01} & \textbf{98.16}\\
\midrule
\multirow{16}{*}{\textbf{LLaVA-OV1.5-8B}} & FP16 & -- & 84.36 & 86.48 & 68.20 & 56.11 & 80.09 & 66.10 & 111.44 & 145.01 & 100.00\\\cmidrule(lr){2-12}
 & \multirow{5}{*}{W4A8} & MBQ~\cite{mbq} & 79.95 & 74.88 & 61.60 & 47.00 & 64.00 & 54.30 & 89.36 & 112.71 & 84.43\\
 &  & QIG~\cite{qig} & 79.27 & 80.96 & 61.07 & 49.56 & 71.26 & 57.02 & 84.86 & 118.35 & 87.31\\
 &  & MASQuant~\cite{hu2026masquant} & 75.19 & 68.48 & 62.67 & 46.44 & 60.06 & 49.05 & 97.79 & 107.74 & 81.78\\
 &  & BFQ~\cite{kang2026beyond} & 80.38 & 81.24 & 61.60 & 49.11 & 73.01 & 58.17 & 92.15 & 121.60 & 89.10\\
 &  & \textbf{Ours} & \textbf{83.58} & \textbf{85.68} & \textbf{66.73} & \textbf{53.44} & \textbf{79.35} & \textbf{64.03} & \textbf{109.75} & \textbf{137.95} & \textbf{97.60}\\
\cmidrule(lr){2-12}
 & \multirow{5}{*}{W4A6} & MBQ~\cite{mbq} & 63.60 & 44.32 & 57.47 & 39.11 & 46.08 & 30.33 & 4.68 & 88.80 & 56.18\\
 &  & QIG~\cite{qig} & 73.02 & 75.52 & 57.73 & 42.00 & 64.26 & 54.86 & 27.49 & 81.42 & 72.18\\
 &  & MASQuant~\cite{hu2026masquant} & 58.32 & 46.88 & 53.07 & 35.89 & 32.84 & 36.11 & 75.13 & 36.07 & 56.63\\
 &  & BFQ~\cite{kang2026beyond} & 76.17 & 76.16 & 56.07 & 41.89 & 68.18 & 56.20 & 78.78 & 91.14 & 79.86\\
 &  & \textbf{Ours} & \textbf{83.35} & \textbf{85.48} & \textbf{66.07} & \textbf{53.22} & \textbf{79.48} & \textbf{63.10} & \textbf{110.16} & \textbf{138.36} & \textbf{97.29}\\
\cmidrule(lr){2-12}
 & \multirow{5}{*}{W4A4} & MBQ~\cite{mbq} & 1.30 & 0.00 & 3.87 & 26.78 & 0.00 & 0.00 & 0.03 & 0.09 & 6.88\\
 &  & QIG~\cite{qig} & 2.20 & 0.00 & 4.80 & 26.33 & 0.00 & 0.00 & 0.04 & 0.09 & 7.08\\
 &  & MASQuant~\cite{hu2026masquant} & 1.55 & 0.00 & 4.53 & 25.22 & 0.00 & 0.00 & 0.04 & 0.09 & 6.69\\
 &  & BFQ~\cite{kang2026beyond} & 1.36 & 0.00 & 4.20 & 25.56 & 0.00 & 0.00 & 0.02 & 0.05 & 6.67\\
 &  & \textbf{Ours} & \textbf{83.16} & \textbf{85.32} & \textbf{64.80} & \textbf{53.56} & \textbf{78.62} & \textbf{62.81} & \textbf{109.00} & \textbf{135.82} & \textbf{96.55}\\
\midrule
\multirow{16}{*}{\textbf{InternVL3.5-8B}} & FP16 & -- & 82.29 & 86.56 & 68.33 & 57.78 & 78.07 & 58.14 & 117.57 & 120.99 & 100.00\\\cmidrule(lr){2-12}
 & \multirow{5}{*}{W4A8} & MBQ~\cite{mbq} & 79.57 & 83.80 & 63.93 & 50.67 & 71.15 & 53.33 & 104.36 & 109.20 & 92.08\\
 &  & QIG~\cite{qig} & 77.82 & 82.88 & 62.13 & 49.33 & 70.88 & 53.16 & 105.38 & 104.36 & 90.59\\
 &  & MASQuant~\cite{hu2026masquant} & 79.08 & 84.40 & 63.47 & 49.33 & 69.62 & 51.26 & 108.31 & 99.67 & 90.46\\
 &  & BFQ~\cite{kang2026beyond} & 78.43 & 83.88 & 62.80 & 53.00 & 70.83 & 55.42 & 106.08 & 103.60 & 92.22\\
 &  & \textbf{Ours} & \textbf{81.38} & \textbf{86.48} & \textbf{66.73} & \textbf{57.56} & \textbf{77.34} & \textbf{58.05} & \textbf{117.32} & \textbf{119.84} & \textbf{99.23}\\
\cmidrule(lr){2-12}
 & \multirow{5}{*}{W4A6} & MBQ~\cite{mbq} & 75.74 & 81.00 & 58.13 & 41.56 & 65.54 & 46.39 & 93.88 & 94.70 & 83.04\\
 &  & QIG~\cite{qig} & 73.09 & 78.36 & 57.33 & 41.44 & 64.31 & 49.82 & 97.73 & 96.80 & 83.25\\
 &  & MASQuant~\cite{hu2026masquant} & 75.94 & 81.88 & 59.53 & 45.22 & 68.13 & 42.19 & 100.66 & 100.25 & 85.05\\
 &  & BFQ~\cite{kang2026beyond} & 74.06 & 79.96 & 56.60 & 40.33 & 66.38 & 52.95 & 98.97 & 99.10 & 84.63\\
 &  & \textbf{Ours} & \textbf{81.06} & \textbf{87.04} & \textbf{67.40} & \textbf{55.67} & \textbf{77.30} & \textbf{59.21} & \textbf{117.13} & \textbf{120.69} & \textbf{99.26}\\
\cmidrule(lr){2-12}
 & \multirow{5}{*}{W4A4} & MBQ~\cite{mbq} & 0.97 & 0.08 & 1.33 & 1.11 & 0.01 & 0.02 & 0.01 & 0.10 & 0.66\\
 &  & QIG~\cite{qig} & 0.45 & 0.00 & 1.07 & 1.78 & 0.00 & 0.00 & 0.01 & 0.03 & 0.65\\
 &  & MASQuant~\cite{hu2026masquant} & 0.49 & 0.00 & 0.00 & 0.22 & 0.00 & 0.00 & 0.01 & 0.02 & 0.13\\
 &  & BFQ~\cite{kang2026beyond} & 0.58 & 0.00 & 0.27 & 1.00 & 0.00 & 0.00 & 0.01 & 0.05 & 0.36\\
 &  & \textbf{Ours} & \textbf{80.28} & \textbf{86.36} & \textbf{65.00} & \textbf{55.56} & \textbf{76.23} & \textbf{57.70} & \textbf{116.45} & \textbf{116.80} & \textbf{97.63}\\
\midrule
\end{tabular}}\end{table*}

\begin{table*}[t]\centering
\caption{Qwen3-VL-32B~\cite{bai2025qwen3} results across 8 multimodal tasks. Ret.: mean FP retention.}\label{tab:main32}
\fitTable{\textwidth}{%
\begin{tabular}{cllrrrrrrrrr}\toprule
Model & Bits & Method & AI2D & ChartQA & MMStar & MMMU & TextVQA & VizWiz & NoCaps & TextCaps & Ret. (\%)\\\midrule
\multirow{16}{*}{\textbf{Qwen3-VL-32B}} & FP16 & -- & 86.72 & 87.32 & 73.13 & 64.22 & 83.70 & 71.21 & 69.80 & 94.31 & 100.00\\\cmidrule(lr){2-12}
 & \multirow{5}{*}{W4A8} & MBQ~\cite{mbq} & 84.68 & 85.28 & 70.87 & 62.33 & 80.25 & 69.02 & 64.62 & 90.38 & 96.31\\
 &  & QIG~\cite{qig} & 84.42 & 83.28 & 70.87 & \textbf{63.44} & 80.03 & 68.27 & 64.35 & 90.69 & 96.03\\
 &  & MASQuant~\cite{hu2026masquant} & 85.01 & 84.32 & 71.33 & 61.44 & 81.18 & 69.28 & 67.55 & 92.18 & 97.07\\
 &  & BFQ~\cite{kang2026beyond} & 84.88 & 85.12 & 70.53 & \textbf{63.44} & 80.65 & 69.22 & 63.87 & 87.59 & 96.07\\
 &  & \textbf{Ours} & \textbf{86.24} & \textbf{87.12} & \textbf{72.07} & 63.11 & \textbf{82.89} & \textbf{71.81} & \textbf{71.39} & \textbf{95.41} & \textbf{99.92}\\
\cmidrule(lr){2-12}
 & \multirow{5}{*}{W4A6} & MBQ~\cite{mbq} & 83.23 & 81.56 & 68.00 & 60.00 & 77.66 & 64.96 & 61.95 & 87.74 & 92.70\\
 & & QIG~\cite{qig} & 79.79 & 74.76 & 63.73 & 49.44 & 74.94 & 63.23 & 58.78 & 81.06 & 86.28\\
 & & MASQuant~\cite{hu2026masquant} & 16.71 & 21.96 & 27.47 & 20.11 & 39.83 & 36.86 & 46.68 & 62.67 & 43.25\\
 & & BFQ~\cite{kang2026beyond} & 81.35 & 79.64 & 66.13 & 57.33 & 76.87 & 62.94 & 60.23 & 82.45 & 89.83\\
 & & \textbf{Ours} & \textbf{86.37} & \textbf{86.84} & \textbf{71.53} & \textbf{64.00} & \textbf{83.09} & \textbf{71.76} & \textbf{71.63} & \textbf{95.08} & \textbf{100.00}\\
\cmidrule(lr){2-12}
 & \multirow{5}{*}{W4A4} & MBQ~\cite{mbq} & 0.65 & 0.04 & 0.47 & 0.67 & 0.00 & 0.00 & 0.01 & 0.04 & 0.32\\
 &  & QIG~\cite{qig} & 1.75 & 0.00 & 1.20 & 2.11 & 0.00 & 0.00 & 0.03 & 0.14 & 0.89\\
 &  & MASQuant~\cite{hu2026masquant} & 0.26 & 0.00 & 0.40 & 0.56 & 0.00 & 0.00 & 0.01 & 0.04 & 0.22\\
 &  & BFQ~\cite{kang2026beyond} & 0.65 & 0.00 & 0.60 & 0.78 & 0.02 & 0.00 & 0.02 & 0.07 & 0.36\\
 &  & \textbf{Ours} & \textbf{85.46} & \textbf{86.08} & \textbf{72.13} & \textbf{62.78} & \textbf{82.40} & \textbf{70.53} & \textbf{71.00} & \textbf{94.08} & \textbf{99.06}\\
\midrule
\end{tabular}}\end{table*}

\begin{table*}[t]
\begin{minipage}[t]{0.48\textwidth}
\vspace{0pt}

\centering
\caption{W4A4 LLM PTQ comparison: 8-task retention (\%) on 8B LVLMs~\cite{bai2025qwen3,an2025llava,wang2025internvl3} and Qwen3-VL generation time. $^\dagger$Atom uses INT8 weights and activations for 128 outlier channels.}
\label{tab:llm_ptq}
\fitTable{\columnwidth}{%
\begin{tabular}{lrrrrr}\toprule
Method & \begin{tabular}[c]{@{}c@{}}Qwen3-\\VL\end{tabular} & \begin{tabular}[c]{@{}c@{}}LLaVA-\\OV1.5\end{tabular} & \begin{tabular}[c]{@{}c@{}}InternVL\\3.5\end{tabular} & Mean & \begin{tabular}[c]{@{}c@{}}Time\\(s)$\downarrow$\end{tabular}\\\midrule
FP (BF16) & 100.00 & 100.00 & 100.00 & 100.00 & 3.75\\\midrule
SmoothQuant~\cite{xiao2023smoothquant} & 1.42 & 7.21 & 0.57 & 3.07 & 2.77\\
OmniQuant~\cite{shao2024omniquant} & 3.43 & 8.30 & 10.70 & 7.48 & 2.96\\
QuaRot~\cite{ashkboos2024quarot} & 93.95 & 90.91 & 95.36 & 93.41 & 3.26\\
DuQuant~\cite{lin2024duquant} & 94.96 & 95.28 & 96.55 & 95.60 & 3.16\\
Atom~\cite{zhao2024atom}$^\dagger$ & 96.79 & 96.29 & 97.25 & 96.78 & 5.45\\
\textbf{Ours} & \textbf{98.16} & \textbf{96.55} & \textbf{97.63} & \textbf{97.45} & \textbf{2.50}\\
\bottomrule\end{tabular}}

\end{minipage}\hfill
\begin{minipage}[t]{0.48\textwidth}
\vspace{0pt}
\centering
\caption{W4A4 component ablation on three LVLMs~\cite{bai2025qwen3,an2025llava,wang2025internvl3}: 4-task retention (\%). H: Hadamard transform; W/A: group quantization for weights/activations. Disabled W/A retain per-channel/per-token quantization.}\label{tab:abl}
\fitTable{\columnwidth}{%
\begin{tabular}{c|ccc|rrrr}\toprule
ID & H & W & A & Qwen3-VL & LLaVA-OV1.5 & InternVL3.5 & Mean\\\midrule
1 & \xmark & \xmark & \xmark & 0.88 & 12.40 & 0.48 & 4.59\\
2 & \xmark & \cmark & \xmark & 1.63 & 12.37 & 0.52 & 4.84\\
3 & \xmark & \xmark & \cmark & 62.89 & 66.86 & 76.19 & 68.65\\
4 & \xmark & \cmark & \cmark & 91.39 & 92.30 & 91.09 & 91.59\\
5 & \cmark & \xmark & \xmark & 85.04 & 85.67 & 85.13 & 85.28\\
6 & \cmark & \cmark & \xmark & 87.56 & 91.03 & 90.61 & 89.73\\
7 & \cmark & \xmark & \cmark & 96.37 & 96.54 & 94.75 & 95.89\\
8 & \cmark & \cmark & \cmark & \textbf{97.18} & \textbf{97.62} & \textbf{98.13} & \textbf{97.64}\\
\bottomrule\end{tabular}}
\end{minipage}
\end{table*}

Using these packed activations, our dedicated decode matrix--vector multiplication (GEMV) kernel computes weight--activation products with group corrections integrated into integer accumulation. To enable this computation, we combine weight encoding inspired by QServe~\cite{lin2025qserve} with token-anchored activation scales:
\begin{equation}
\hat w_{ok}=c_o(m_{og}u_{ok}+b_{og}),\quad
\hat x_{tk}=\alpha_t\mu_{tg}q_{tk}.
\label{eq:integer_representation}
\end{equation}
Here, $t,o,k$ index tokens, output channels, and input channels ($k\in g$). Weight/activation codes $u_{ok},q_{tk}$ use integer multipliers $m_{og},\mu_{tg}$ and weight offsets $b_{og}$. Channel/token scales $c_o,\alpha_t$ are shared across groups. To integrate group-wise corrections into integer accumulation, we compute
\begin{equation}
P_{tog}=\sum_{k\in g}q_{tk}u_{ok},\qquad
S_{tg}=\sum_{k\in g}q_{tk}.
\label{eq:group_sums}
\end{equation}
$P_{tog}$ denotes the code dot product and $S_{tg}$ the activation-code sum. Group correction and output reconstruction give
\begin{equation}
\begin{aligned}
A_{tog}&=\mu_{tg}(m_{og}P_{tog}+b_{og}S_{tg}),\\
\hat y_{to}&=\sum_k\hat x_{tk}\hat w_{ok}=\alpha_tc_o\sum_g A_{tog}.
\end{aligned}
\label{eq:integer}
\end{equation}
Here, $A_{tog}$ is the corrected group contribution and $\hat y_{to}$ the reconstructed output. Corrections are fused with integer accumulation before applying the common scale. Further implementation details and additional runtime optimizations are described in the supplementary material.

\section{Experiments}
\subsection{Experimental setup}
\label{sec:setup}
\paragraph{Implementation details.}
We randomly sample 128 COCO~\cite{lin2014microsoft} image--caption pairs from ShareGPT4V~\cite{chen2024sharegpt4v} to calibrate all methods. We set Hadamard block size $h=32$ and quantization group size $G=128$ for our method. All experiments use an NVIDIA RTX A6000 GPU, except 32B experiments on an RTX PRO 6000.

\paragraph{Datasets, models, and baselines.}
We evaluate Qwen3-VL-8B/32B~\cite{bai2025qwen3}, LLaVA-OneVision-1.5-8B~\cite{an2025llava}, and InternVL3.5-8B~\cite{wang2025internvl3} on question answering (QA) tasks (AI2D~\cite{kembhavi2016diagram}, ChartQA~\cite{chartqa}, MMStar~\cite{chen2024we}, MMMU~\cite{mmmu}, TextVQA~\cite{singh2019towards}, and VizWiz~\cite{vizwiz}) and open-ended generation tasks (NoCaps~\cite{agrawal2019nocaps} and TextCaps~\cite{sidorov2020textcaps}). We report task scores and mean full-precision (FP) performance retention (\%). Main results use all 8 tasks; ablations use AI2D, ChartQA, MMMU, and NoCaps. We compare with MBQ~\cite{mbq}, QIG~\cite{qig}, MASQuant~\cite{hu2026masquant}, and BFQ~\cite{kang2026beyond} using their official implementations and calibration policies.

\subsection{Comparison with existing approaches}
\Cref{tab:main8} shows post-training quantization (PTQ) results on three 8B large vision-language models (LVLMs) across 8 tasks. Our method substantially outperforms state-of-the-art LVLM PTQ methods across all evaluated quantization settings. In particular, at W4A4, the compared methods suffer near-total performance collapse, retaining less than 8\% of FP performance, whereas ours preserves 96.55--98.16\% across the three models. Furthermore, \Cref{tab:main32} shows even higher mean retention on Qwen3-VL-32B, exceeding 99\% across all evaluated quantization settings.

\begin{table*}[t]
\centering\includegraphics[width=\textwidth]{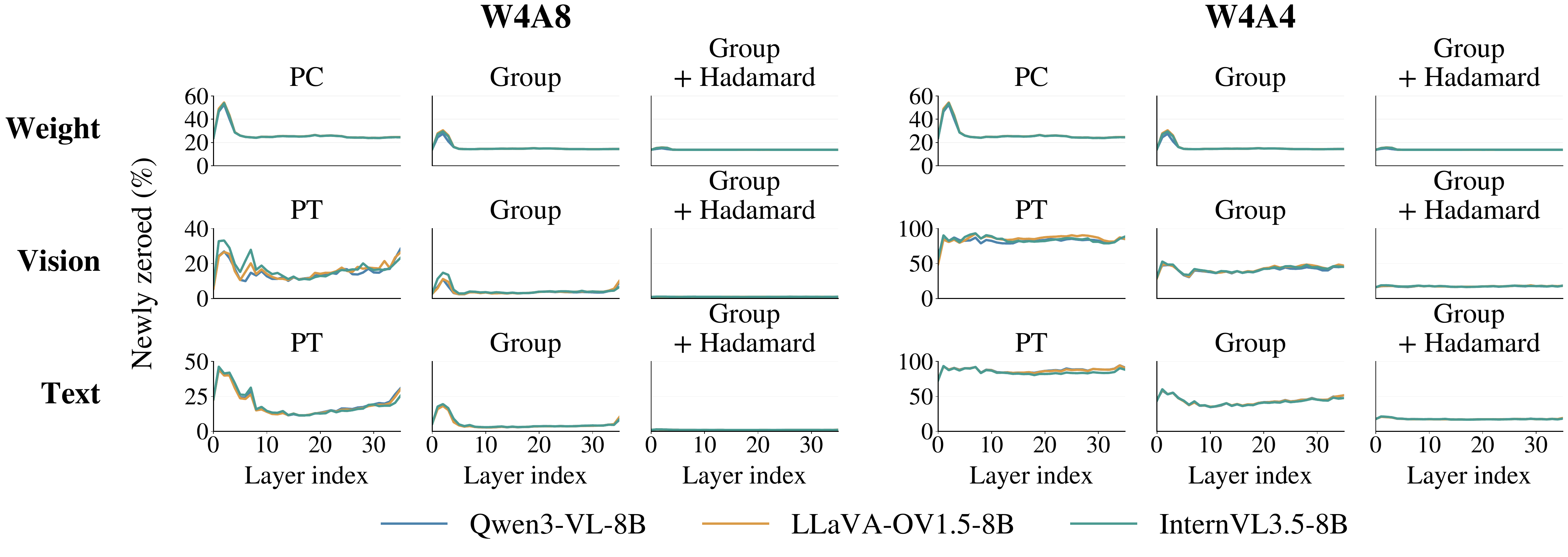}
\captionof{figure}{Percentages of nonzero weights and activations mapped to zero by quantization in each layer at W4A8 (left) and W4A4 (right).}
\label{fig:zero}
\end{table*}

Rotation and group quantization are widely used to mitigate outliers in LLM PTQ. To assess their effectiveness in LVLMs, we compare our method with SmoothQuant~\cite{xiao2023smoothquant} and OmniQuant~\cite{shao2024omniquant} as scaling-based methods, rotation-based QuaRot~\cite{ashkboos2024quarot} and DuQuant~\cite{lin2024duquant}, and Atom~\cite{zhao2024atom}, which combines group quantization with mixed-precision outlier handling. For fair speed comparisons, all methods retain their quantization policies and official low-bit kernels but run within our optimized runtime with the same output length. Timing is measured on Qwen3-VL-8B on 25 samples each from AI2D and NoCaps. \Cref{tab:llm_ptq} shows substantially higher W4A4 retention for rotation- and group-based methods than the LVLM PTQ baselines in \cref{tab:main8}. However, their speed gains remain limited despite the optimized runtime, and mixed-precision Atom is slower than FP. Our method achieves the highest mean retention while reducing generation time by 15.3--54.0\% relative to these baselines.

\subsection{Ablation study}
We evaluate how weight grouping, activation grouping, and Hadamard transforms affect performance retention across three LVLMs~\cite{bai2025qwen3,an2025llava,wang2025internvl3}. \Cref{tab:abl} shows that all three components substantially improve performance. Since per-channel (PC) weight and per-token (PT) activation quantization causes severe information loss in both weights and activations, redistributing values with Hadamard transforms alone yields substantial recovery. Without Hadamard transforms, strong performance retention is achieved only when both weights and activations are group-quantized. Combining all three components achieves the best performance across models.

\begin{figure}[t]
\centering
\includegraphics[width=\columnwidth]{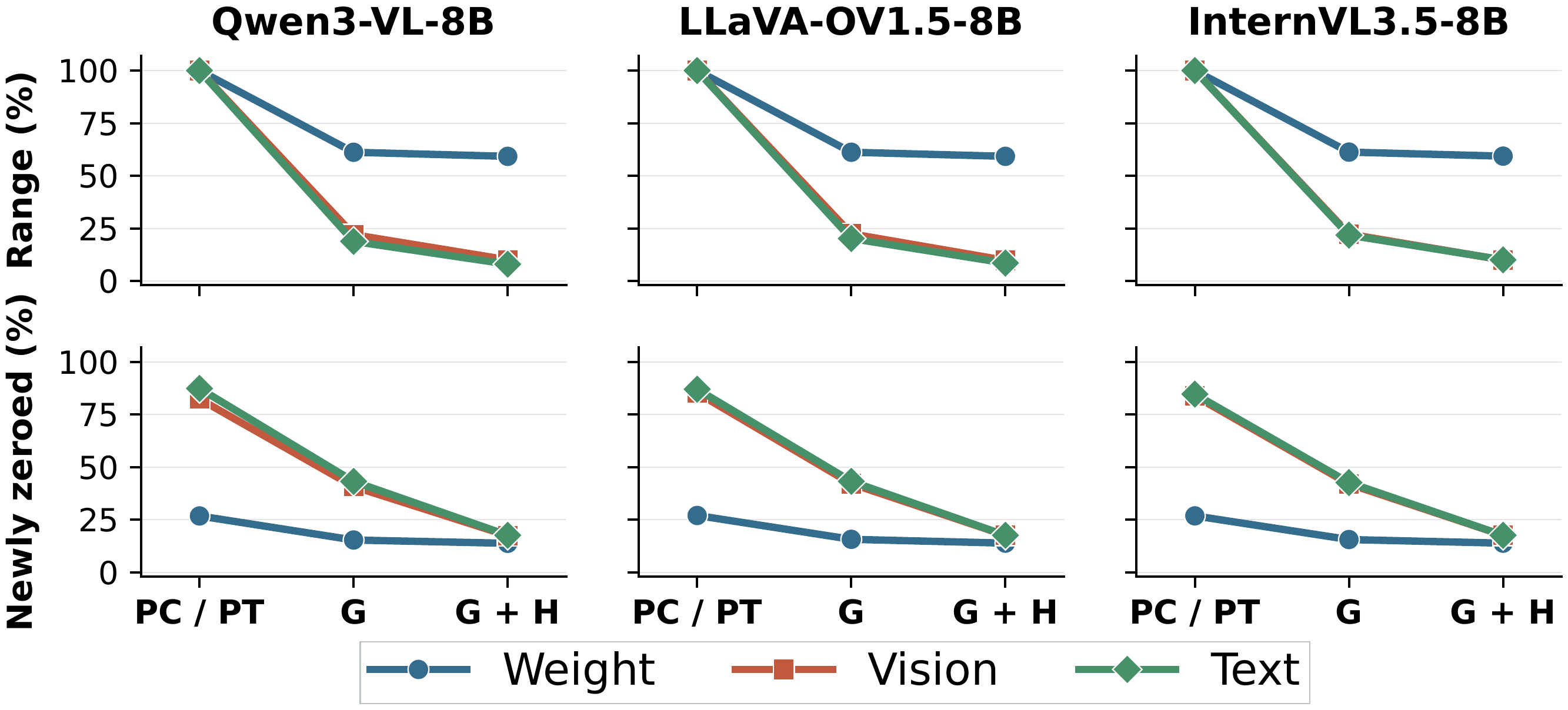}
\caption{\textbf{Quantization ranges and newly-zeroed values at W4A4.} Ranges relative to PC/PT (top) and percentages of nonzero values quantized to zero (bottom) for weights and vision/text activations. G: group quantization; H: Hadamard transform.}
\label{fig:range}

\end{figure}

\begin{figure}[t]
\centering\includegraphics[width=\columnwidth]{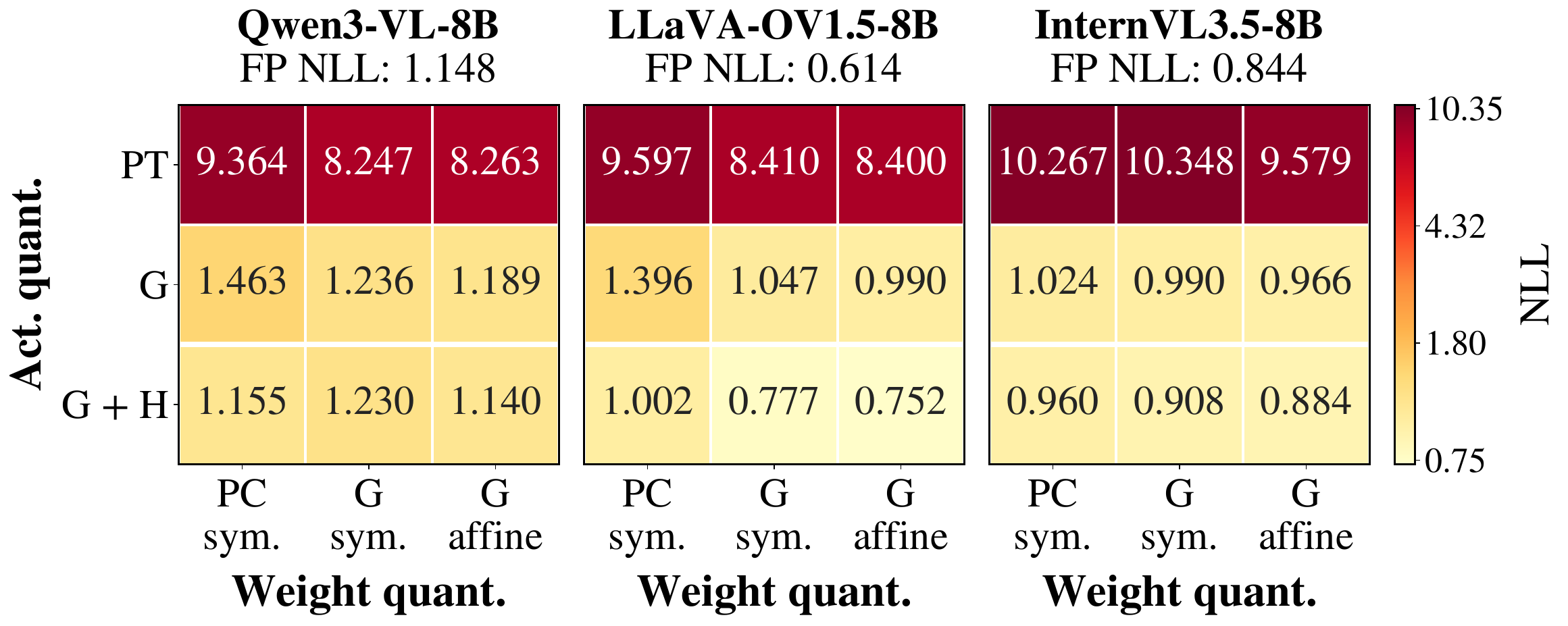}
\caption{Token-weighted negative log-likelihood (NLL) on 128 calibration examples at W4A4. Columns vary weight representation and rows vary activation representation. The Hadamard row transforms both operands.}
\label{fig:nll}

\end{figure}

Furthermore, we investigate how our method reduces quantization-induced information loss throughout the model. We collect layerwise weight distributions and activations on 32 image--caption pairs randomly drawn from the same dataset, excluding calibration samples. \Cref{fig:zero} shows the fraction of nonzero weights and activations mapped to zero by quantization in each layer. At W4A8, activations in early layers show high zero-mapping rates despite their 8-bit precision, and weights also exhibit pronounced zero mapping in these layers. At W4A4, a large fraction of activation values is mapped to zero across all layers, confirming that the outlier-induced information loss identified in \cref{sec:motivation} occurs throughout the models. Across both settings, group quantization reduces zero mapping, while adding Hadamard transforms keeps it consistently low across layers. These reductions demonstrate that our method effectively mitigates outliers and limits quantization-induced information loss.

\Cref{fig:range} summarizes quantization ranges and zero-mapping rates across decoder weights and vision/text activations at W4A4 on the same samples. Group quantization narrows the ranges and reduces zero mapping, with further reductions from Hadamard transforms.

To evaluate the resulting improvement in prediction quality, \Cref{fig:nll} compares response-token negative log-likelihood (NLL) on the 128 calibration image--caption pairs. Combining group quantization and Hadamard transforms reduces NLL from 9.36--10.27 under PC/PT to 0.75--1.14, approaching the corresponding FP levels of 0.61--1.15. The near-FP NLL values demonstrate a substantial reduction in quantization-induced prediction error.

To assess whether more complex calibration offers benefits over uniform search, we replace our uniform channel-wise equalization (CWE) scale search with alternative calibration strategies, keeping all other settings unchanged. \Cref{tab:matched} shows that uniform search matches or slightly exceeds their mean retention. These results suggest that once quantization error is reduced, complex calibration strategies provide no consistent performance gains while increasing calibration cost.

\begin{table}[t]\centering
\caption{Calibration strategy comparison at W4A4: 4-task retention (\%) on three LVLMs~\cite{bai2025qwen3,an2025llava,wang2025internvl3}. Only the calibration strategy changes; all other settings are identical to our method.}\label{tab:matched}
\fitTable{\columnwidth}{%
\begin{tabular}{lrrrr}\toprule
Method & Qwen3-VL & LLaVA-OV1.5 & InternVL3.5 & Mean\\\midrule
MBQ~\cite{mbq} & 96.62 & \textbf{99.14} & 96.69 & 97.48\\
QIG~\cite{qig} & 96.23 & 97.84 & \textbf{98.22} & 97.43\\
BFQ~\cite{kang2026beyond} & 96.32 & 97.45 & 98.04 & 97.27\\
Uniform & \textbf{97.18} & 97.62 & 98.13 & \textbf{97.64}\\
\bottomrule\end{tabular}}\end{table}

\begin{table}[t]\centering
\caption{W4A4 group-size (G) and Hadamard-size (H) ablations. Retention averages 4 tasks over three models~\cite{bai2025qwen3,an2025llava,wang2025internvl3}. G varies at H32 and H at G128. Costs are measured on Qwen3-VL-8B.}
\label{tab:group_hadamard_size}
\fitTable{\columnwidth}{%
\begin{tabular}{lrrr}\toprule
Metric & G64 & G128 & G256\\\midrule
Mean retention (\%)$\uparrow$ & 98.30 & 97.64 & 97.04\\
Prefill (ms)$\downarrow$ & 149.15 & 114.28 & 105.73\\
Peak allocated (GiB)$\downarrow$ & 7.332 & 7.232 & 7.181\\\midrule
Metric & H16 & H32 & H64\\\midrule
Mean retention (\%)$\uparrow$ & 97.41 & 97.64 & 97.13\\
\bottomrule\end{tabular}}\end{table}

We also evaluate the choice of group and Hadamard block sizes. \Cref{tab:group_hadamard_size} shows that smaller groups improve mean retention but require scale computation for more groups, particularly increasing prefill cost. We therefore select G128 to balance accuracy and cost, and select H32 for its highest mean retention.

\subsection{Cost Analysis}

\begin{table}[t]\centering
\caption{W4A4 full-calibration wall time on 8B models~\cite{bai2025qwen3,an2025llava,wang2025internvl3}.}\label{tab:calmain}
\fitTable{\columnwidth}{%
\begin{tabular}{lrrr}\toprule
& \multicolumn{3}{c}{Calibration time (min)}\\
\cmidrule(lr){2-4}
Method & Qwen3-VL & LLaVA-OV1.5 & InternVL3.5\\\midrule
MBQ~\cite{mbq} & 39.96 & 48.18 & 250.76\\
QIG~\cite{qig} & 73.82 & 89.84 & 432.41\\
BFQ~\cite{kang2026beyond} & 52.90 & 62.33 & 246.06\\
MASQuant~\cite{hu2026masquant} & 260.04 & 260.73 & 294.95\\
\textbf{Ours} & \textbf{23.88} & \textbf{27.37} & \textbf{130.30}\\
\bottomrule\end{tabular}}\end{table}

\begin{table}[t]\centering
\caption{W4A4 inference costs on Qwen3-VL-8B~\cite{bai2025qwen3}. Parentheses show changes from BF16. Group+H is a direct implementation without the proposed execution optimizations. PC/PT and Ours use optimized execution.}
\label{tab:evalmain}
\fitTable{\columnwidth}{%
\begin{tabular}{lrr}\toprule
Method & Generation time (s)$\downarrow$ & Peak allocated (GiB)$\downarrow$\\\midrule
FP (BF16) & 3.75 & 16.86\\
Group+H & 11.86 ($\uparrow$216.30\%) & 7.29 ($\downarrow$56.77\%)\\
PC/PT & 2.42 ($\downarrow$35.31\%) & 7.13 ($\downarrow$57.72\%)\\
Ours & 2.50 ($\downarrow$33.18\%) & 7.23 ($\downarrow$57.12\%)\\
\bottomrule\end{tabular}}\end{table}

To evaluate the efficiency of our pipeline, we compare the calibration cost of our uniform CWE scale search with other LVLM PTQ methods on the same 128 calibration samples. \Cref{tab:calmain} reports the total elapsed time at W4A4, including loading, preprocessing, calibration, and packing. Our method reduces calibration time by approximately 40--91\% relative to LVLM PTQ baselines across the three models, avoiding their additional modality-sensitivity analysis. These results highlight the advantages of our method in accuracy, simplicity, and calibration efficiency.

We then evaluate whether the execution optimizations preserve the efficiency benefits of low-bit quantization. \Cref{tab:evalmain} compares generation time and peak allocated memory on Qwen3-VL-8B at W4A4, using 25 randomly selected samples each from AI2D and NoCaps. Group + Hadamard denotes a direct implementation of local Hadamard transforms and group quantization without our proposed group-aware execution optimizations. For a fair comparison, PC/PT and our method run under the same optimized runtime settings. Although the direct Group + Hadamard implementation reduces memory, it increases generation time relative to full precision. Our optimized execution instead reduces generation time by 33.18\% and peak memory by 57.12\%, achieving savings comparable to optimized per-channel weight and per-token activation quantization. Additional measurement details, results, and code for reproducing our experiments are included in the supplementary material.

\section{Conclusion}
In this paper, we show that outliers severely limit the precision of per-channel weight and per-token activation quantization in LVLMs. To address this issue, we introduce Rotate-and-Refine Quantization, combining local Hadamard transforms and group quantization with efficient group-aware execution. Across 8 multimodal tasks and 8B/32B models, it substantially improves performance retention with simple calibration with time and memory savings. Our results emphasize accurate quantization representations and efficient execution over increasingly complex calibration.

{\small\bibliographystyle{ieeenat_fullname}\bibliography{main}}
\makeatletter
\setlength{\@fptop}{0pt}
\setlength{\@dblfptop}{0pt}
\setlength{\@fpbot}{0pt plus 1fil}
\setlength{\@dblfpbot}{0pt plus 1fil}
\makeatother
\twocolumn[\centering\Large\bfseries Supplementary Material\par\vspace{1em}]
\raggedbottom
\renewcommand{\theHsection}{supp.\arabic{section}}
\renewcommand{\theHfigure}{supp.\arabic{figure}}
\renewcommand{\theHtable}{supp.\arabic{table}}
\renewcommand{\theHequation}{supp.\arabic{equation}}
\setcounter{figure}{0}\renewcommand{\thefigure}{S\arabic{figure}}
\setcounter{table}{0}\renewcommand{\thetable}{S\arabic{table}}
\setcounter{equation}{0}\renewcommand{\theequation}{S\arabic{equation}}
\setcounter{section}{0}\renewcommand{\thesection}{\Alph{section}}

\section{Quantization and runtime optimizations}
\label{sec:supp_implementation}
This section provides implementation details supporting our quantization and efficient execution. We describe the calibration procedure, the weight and activation representations that enable integer group-wise computation, and the runtime optimizations used for inference.

\subsection{Uniform scale search}
For channel-wise equalization (CWE), we search 20 uniformly spaced exponents to form candidate scales:
\begin{equation}
\begin{aligned}
r_j&=j/19,\qquad s_k=\max(a_k^{r_j},10^{-4}),\\
s_k&\leftarrow s_k/\sqrt{(\max_l s_l)(\min_l s_l)}.
\end{aligned}
\end{equation}
Here $j\in\{0,\ldots,19\}$, $a_k$ is the mean absolute calibration activation in input channel $k$, and $s_k$ is its equalization scale. Each candidate is evaluated with equalization, Hadamard transformation, and group quantization, including the weight refitting and activation multiplier constraints described below. This evaluates candidates under the representation used at inference. We then select the candidate with the lowest mean absolute output reconstruction error.

\subsection{Weight representation and offline refitting}
\paragraph{Weight representation.}
We use the QServe-compatible representation~\cite{lin2025qserve} $\hat w_{ogj}=c_o(m_{og}q_{ogj}+b_{og})$, where $o$ indexes output channels, $g$ indexes input-channel groups, and $j$ indexes entries within a group. The four-bit codes satisfy $q_{ogj}\in\{0,\ldots,15\}$, while the integer coefficients satisfy $m_{og}\in\{1,\ldots,17\}$ and $-128\le b_{og}\le127-15m_{og}$. These bounds keep every expanded codebook value within signed INT8. Inference stores packed four-bit codes, UINT8 multipliers, INT8 offsets, and FP16 row scales $c_o$.

\paragraph{Offline refitting.}
The target $V_{ogj}$ is the weight reconstructed from affine group quantization after equalization and Hadamard transformation. A row-wise symmetric INT8 quantizer supplies $c_o$, which remains fixed during refitting. For each group, we minimize the normalized weight reconstruction error under the constraints above:
\begin{equation}
E_{og}=\sum_j\left(m_{og}q_{ogj}+b_{og}-\frac{V_{ogj}}{c_o}\right)^2.
\end{equation}
Starting from a range-based initialization, we perform three alternating updates: fit integer coefficients by rounding and constraining the least-squares solution for the current codes, then reassign the four-bit codes. Each group accepts an update only when its error decreases. Refitting is included in calibration-candidate evaluation; the selected codes and coefficients are packed offline, with no iterative refitting during inference.

\subsection{Token-anchored activation scales}
For transformed activation $x_{tk}$ at token $t$, the symmetric quantization scale of group $g$, before imposing integer-multiplier constraints, is
\begin{equation}
s_{tg}=\max_{k\in g}|x_{tk}|/(2^{b_a-1}-1),
\end{equation}
where $b_a\in\{4,6,8\}$ is the activation bitwidth. To express these group scales as integer multiples of a common token scale, we round $\max_g s_{tg}/8$ upward to an FP16 token scale $\alpha_t$ and construct integer group multipliers:
\begin{equation}
\begin{aligned}
\mu_{tg}&=\operatorname{clip}(\lceil s_{tg}/\alpha_t\rceil,1,8),\\
q_{tk}&=\operatorname{clip}\!\left(\operatorname{round}\frac{x_{tk}}{\alpha_t\mu_{tg}},-2^{b_a-1},2^{b_a-1}-1\right),\\
\hat x_{tk}&=\alpha_t\mu_{tg}q_{tk}.
\end{aligned}
\end{equation}
Here $q_{tk}$ is the signed activation code and $\hat x_{tk}$ its reconstruction. Activation quantization is symmetric, without additional learned clipping. We cap the integer group multiplier at 8 to balance quantization accuracy and the range of INT32 accumulation, and use this construction across all evaluated activation bitwidths.

\subsection{Execution and runtime optimizations}
Our implementation combines general runtime optimizations with RRQ-specific group-aware execution. The general optimizations include fusion of root mean square normalization (RMSNorm)~\cite{zhang2019root} and sigmoid-weighted linear unit (SiLU)~\cite{elfwing2018sigmoid} operations where applicable, input reuse across compatible projections, and CUDA Graphs for decoding. RRQ-specific execution instead fuses Hadamard transformation, group activation quantization, and packing, and integrates group-wise corrections into integer accumulation with input-shape-specific execution paths. These mechanisms address the additional transformation and group-scale handling introduced by RRQ.

The direct Group+Hadamard implementation includes weight packing and input reuse but omits our group-aware execution optimizations. Table~\ref{tab:eval_all_models} compares complete execution configurations; it does not isolate the contribution of the fused preprocessing kernel from the other optimizations. Isolating that contribution requires holding the general runtime settings fixed and varying the RRQ-specific components separately.

\section{Details of Distribution Analyses}
\label{sec:supp_diagnostics}
This section provides distribution-selection and measurement details for Figures~2, 3, 5, and~6 of the main paper.

\paragraph{Illustrative distributions in Figures 2 and 3.}
For the examples in Figures~2 and~3 of the main paper, we use negative log-likelihood (NLL) recovery from the projection-restoration experiment to rank layers. For the illustrative distributions, we select a vector with the smallest fraction of values exceeding 10\% of its maximum absolute value. This criterion highlights distributions in which a few large values dominate the range; their prevalence is assessed separately through the layerwise and aggregate analyses below. Recovery values depend on interactions between components and should not be summed to partition the total quantization error.

\paragraph{Layerwise and aggregate statistics in Figures 5 and 6.}
For the layerwise and aggregate analyses in Figures~5 and~6 of the main paper, we collect statistics over decoder weights and activation inputs, counting inputs shared by multiple projections only once. We obtain activation inputs from the full-precision (FP) model so that errors from preceding quantized layers do not affect the comparison. After excluding padding, we measure zero mapping as the fraction of nonzero values mapped to zero by quantization. Because Hadamard transformation mixes features, we compute statistics for each resulting distribution rather than track individual values through the transformation.

\section{Impact of Integer-Compatible Quantization}
\label{sec:supp_controls}
We next assess whether the integer-compatible representation needed for efficient execution compromises accuracy. Table~\ref{tab:free} compares it with native group quantization for weights, activations, or both. Each configuration uses Hadamard transforms and is independently calibrated by uniform scale search on the same 128 calibration pairs. We report mean retention over AI2D, ChartQA, MMMU, and NoCaps.

\begin{table}[t]\centering
\small\setlength{\tabcolsep}{2pt}
\caption{W4A4 performance retention (\%) with our integer-compatible (I) and native (N) group quantization on three 8B models~\cite{bai2025qwen3,an2025llava,wang2025internvl3}. Mean averages the three model results.}\label{tab:free}
\begin{tabular}{llrrrr}\toprule
Weight&Act.&\shortstack{Qwen3-\\VL}&\shortstack{LLaVA-\\OV1.5}&\shortstack{InternVL\\3.5}&Mean\\\midrule
N&N&96.47&97.41&98.22&97.37\\
I&N&96.85&98.23&97.22&97.43\\
N&I&97.43&97.46&98.19&97.69\\\midrule
I&I (Ours)&97.18&97.62&98.13&97.64\\\bottomrule
\end{tabular}\end{table}
\begin{table}[!t]\centering
\caption{W4A4 inference costs on three 8B LVLMs~\cite{bai2025qwen3,an2025llava,wang2025internvl3}. Parentheses show changes from BF16. Group+H is a direct implementation without the proposed execution optimizations. PC/PT and Ours use optimized execution.}
\label{tab:eval_all_models}\label{tab:cost4}
\resizebox{\columnwidth}{!}{%
\begin{tabular}{lrr}\toprule
Method & Generation time (s)$\downarrow$ & Peak allocated (GiB)$\downarrow$\\\midrule
\multicolumn{3}{c}{Qwen3-VL-8B}\\\midrule
FP (BF16) & 3.748 & 16.864\\
Group+H & 11.855 ($\uparrow$216.30\%) & 7.291 ($\downarrow$56.77\%)\\
PC/PT & 2.425 ($\downarrow$35.31\%) & 7.130 ($\downarrow$57.72\%)\\
Ours & 2.504 ($\downarrow$33.18\%) & 7.232 ($\downarrow$57.12\%)\\
\midrule
\multicolumn{3}{c}{LLaVA-OV1.5-8B}\\\midrule
FP (BF16) & 4.003 & 16.788\\
Group+H & 11.555 ($\uparrow$188.66\%) & 7.220 ($\downarrow$56.99\%)\\
PC/PT & 2.668 ($\downarrow$33.35\%) & 7.079 ($\downarrow$57.83\%)\\
Ours & 2.720 ($\downarrow$32.05\%) & 7.183 ($\downarrow$57.21\%)\\
\midrule
\multicolumn{3}{c}{InternVL3.5-8B}\\\midrule
FP (BF16) & 3.928 & 16.404\\
Group+H & 12.645 ($\uparrow$221.92\%) & 6.864 ($\downarrow$58.16\%)\\
PC/PT & 2.586 ($\downarrow$34.16\%) & 6.697 ($\downarrow$59.17\%)\\
Ours & 2.641 ($\downarrow$32.76\%) & 6.801 ($\downarrow$58.54\%)\\
\bottomrule\end{tabular}}\end{table}

Our integer-compatible representation achieves 97.64\% mean retention, compared with 97.37\% for native group quantization of both weights and activations. For each of the three configurations containing native group quantization, paired-bootstrap intervals for retention differences relative to ours include zero in all models. These comparisons show no clear accuracy advantage for either representation. Together with the inference results in Section~\ref{sec:supp_cost}, the observed retention supports the use of our integer-compatible representation to improve execution speed while maintaining accuracy.

\begin{figure*}[!t]
\centering
\includegraphics[width=\textwidth,height=0.78\textheight,keepaspectratio]{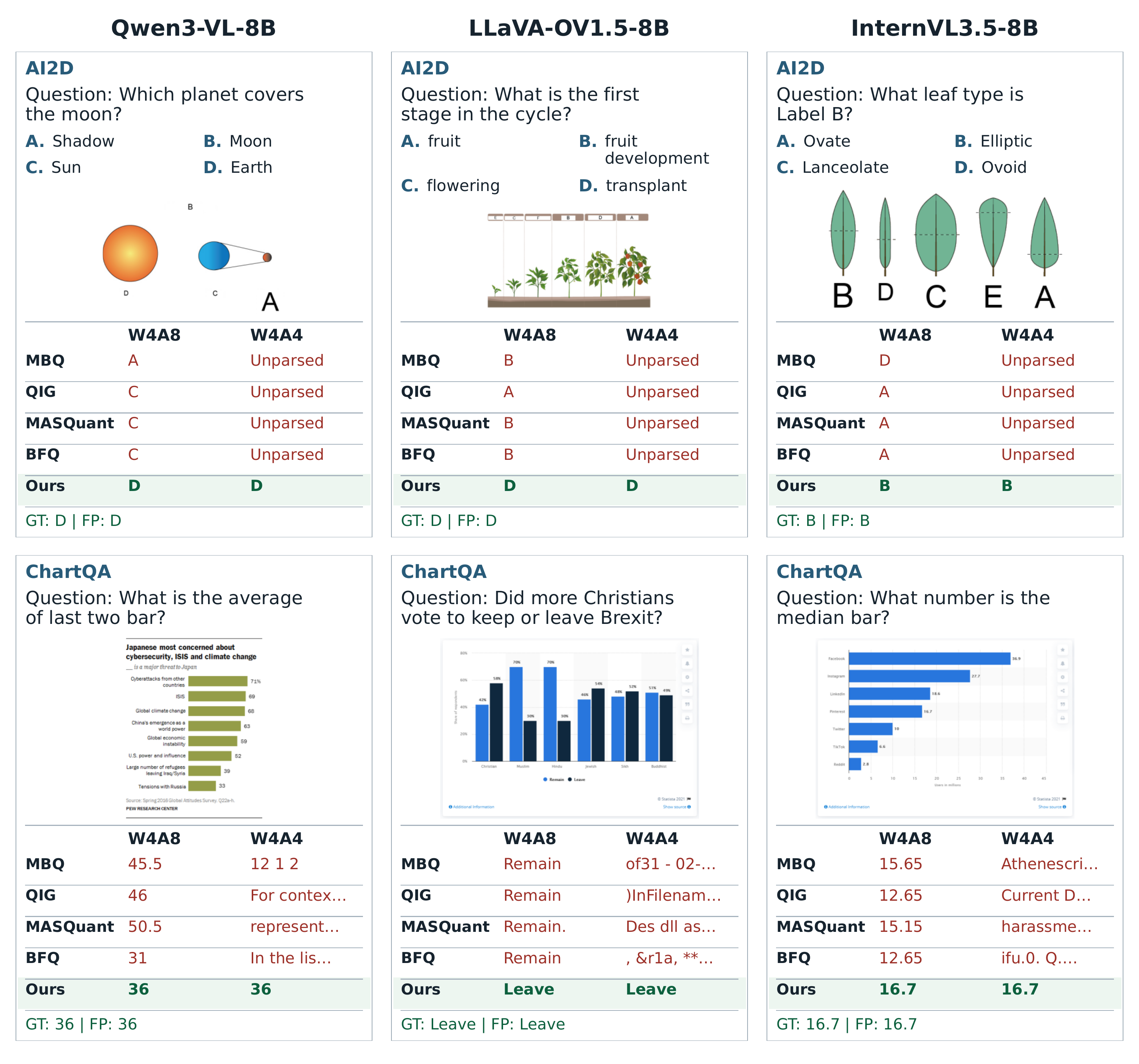}
\caption{Qualitative results on AI2D~\cite{kembhavi2016diagram} and ChartQA~\cite{chartqa}.}
\label{fig:qual_ai2d_chartqa}
\end{figure*}

\begin{figure*}[!t]
\centering
\includegraphics[width=\textwidth,height=0.88\textheight,keepaspectratio]{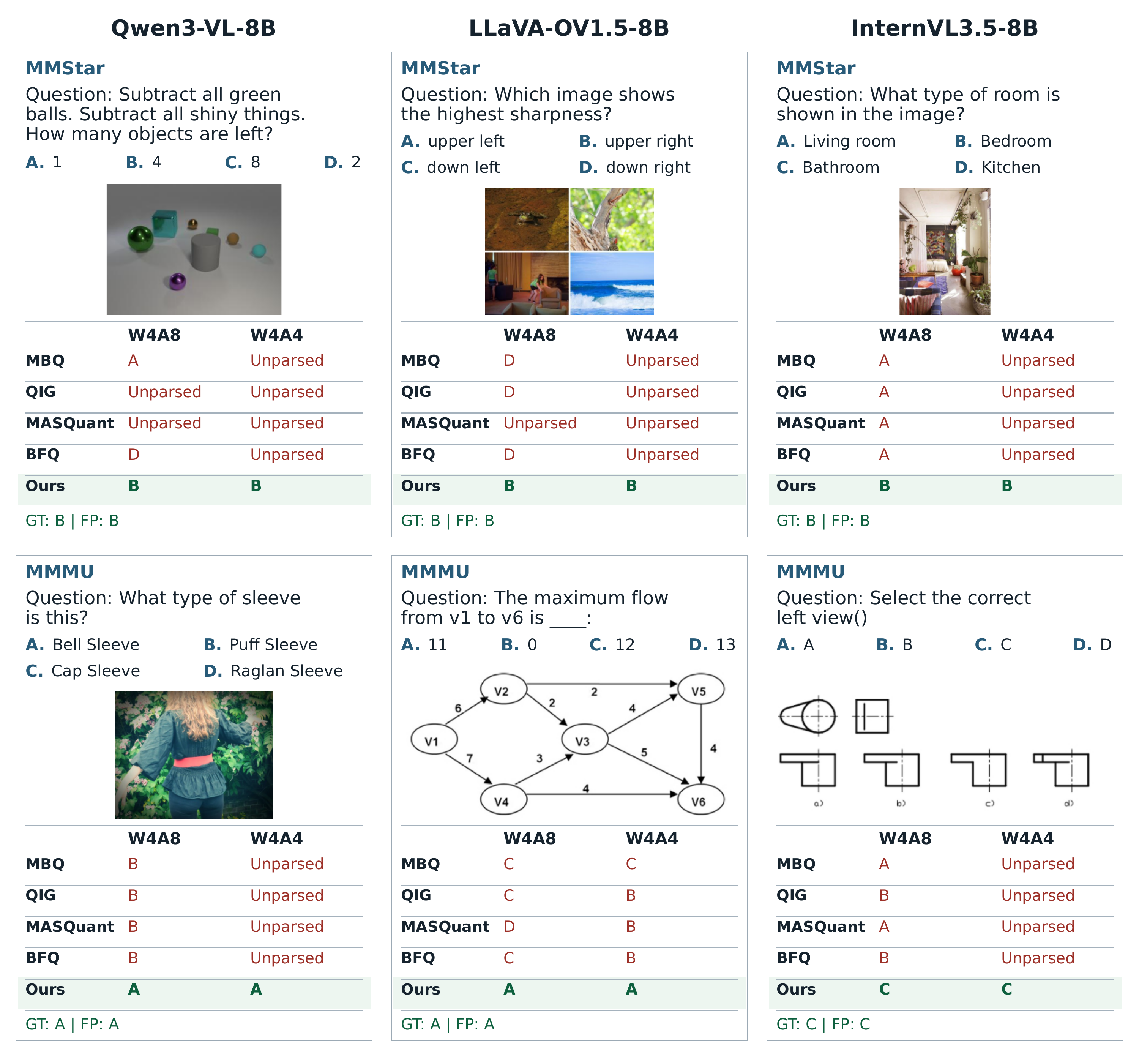}
\caption{Qualitative results on MMStar~\cite{chen2024we} and MMMU~\cite{mmmu}.}
\label{fig:qual_mmstar_mmmu}
\end{figure*}

\begin{figure*}[!t]
\centering
\includegraphics[width=\textwidth,height=0.88\textheight,keepaspectratio]{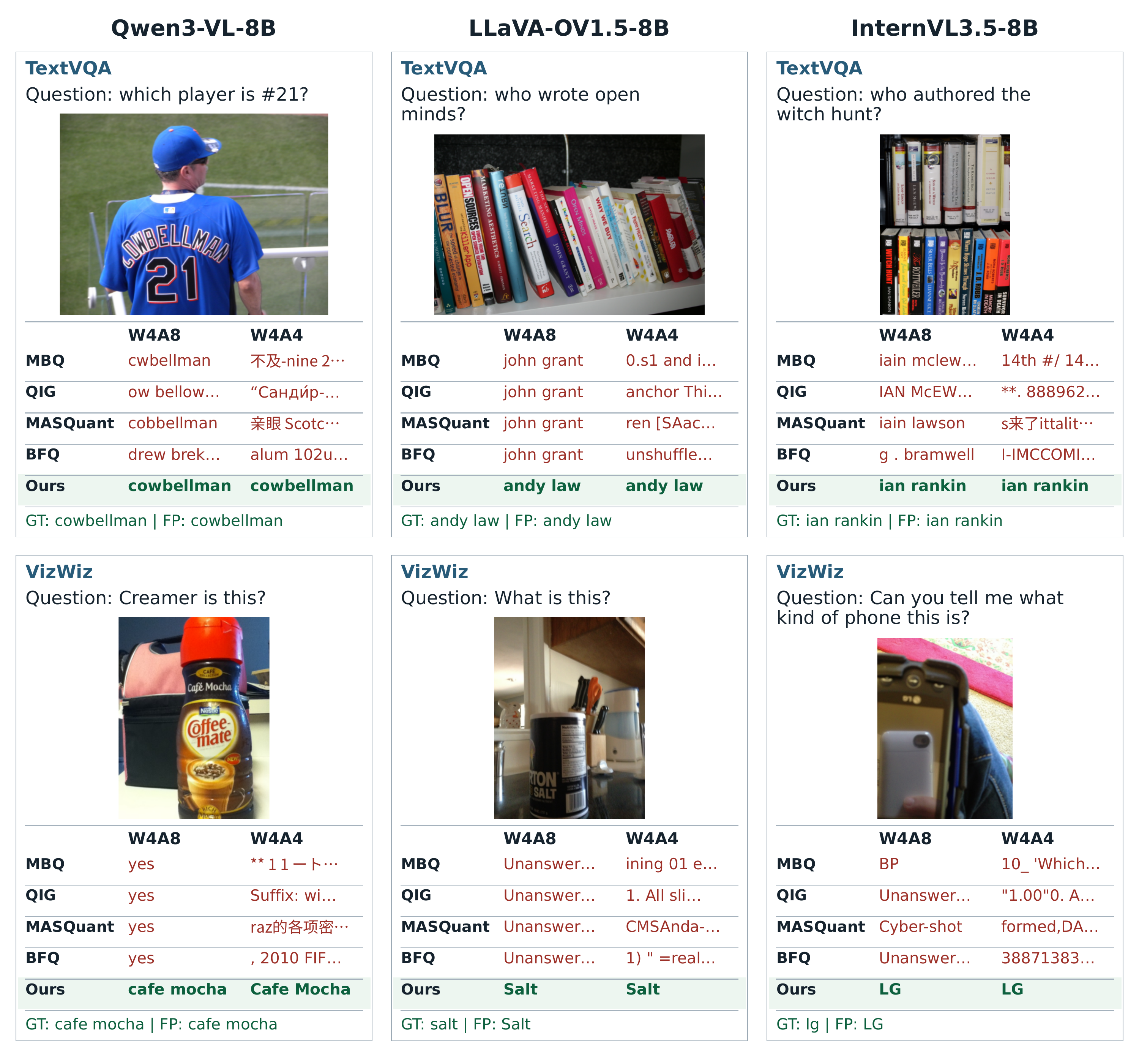}
\caption{Qualitative results on TextVQA~\cite{singh2019towards} and VizWiz~\cite{vizwiz}.}
\label{fig:qual_textvqa_vizwiz}
\end{figure*}

\begin{figure*}[!t]
\centering
\includegraphics[width=\textwidth,height=0.95\textheight,keepaspectratio]{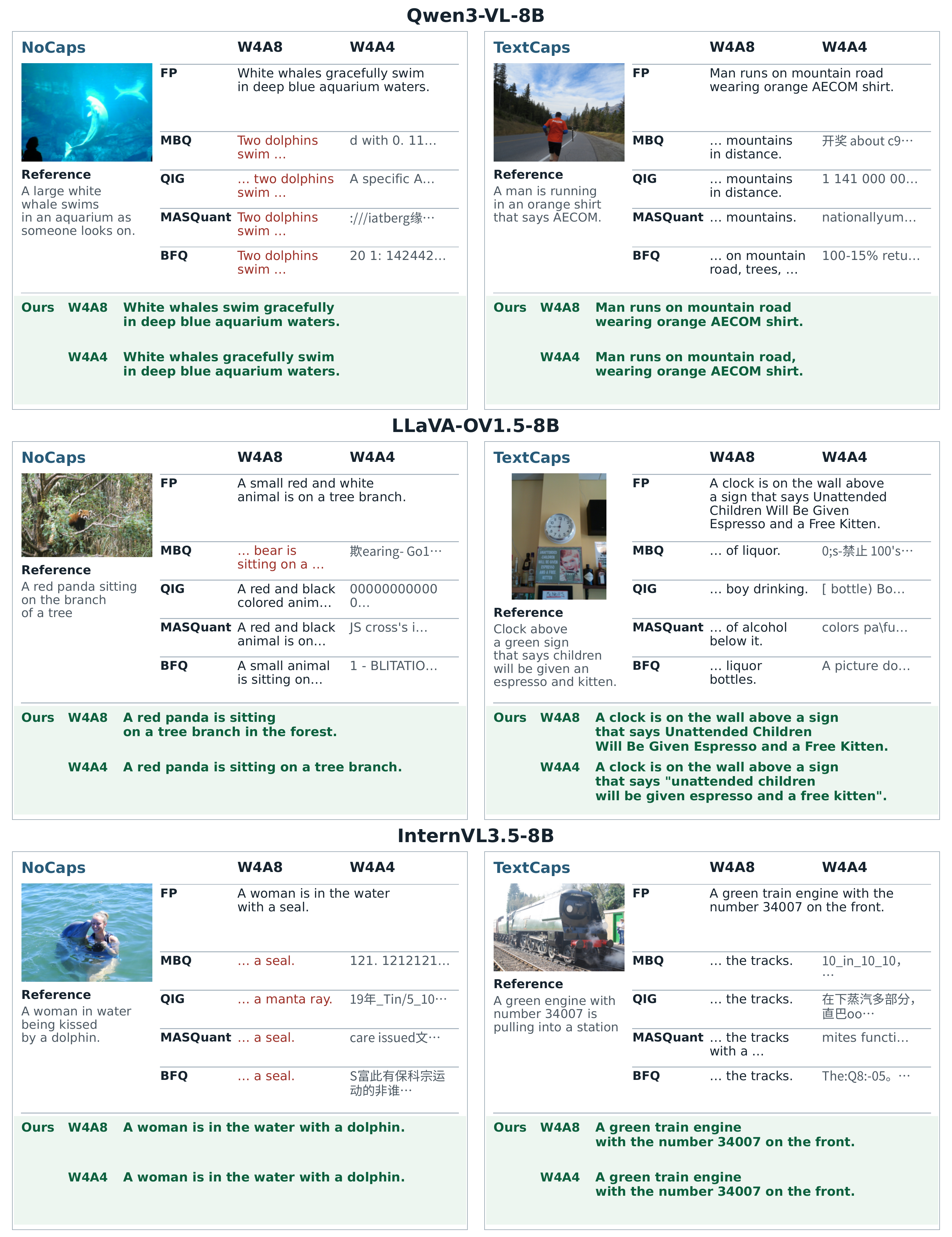}
\caption{Qualitative results on NoCaps~\cite{agrawal2019nocaps} and TextCaps~\cite{sidorov2020textcaps}.}
\label{fig:qual_nocaps_textcaps}
\end{figure*}

\section{Cost measurement and additional results}
\label{sec:supp_cost}
\subsection{Calibration cost}
Following the procedure proposed by its authors, we measure BFQ~\cite{kang2026beyond} calibration time with component analysis parallelized across 3 GPUs, including subsequent fitting. MASQuant~\cite{hu2026masquant} runs compensation fitting over 128 samples for 2 epochs, following its official implementation and recommended settings.

\subsection{Inference cost}
We measure inference on one RTX A6000 GPU with batch size 1, evaluating the same 25 samples each from AI2D and NoCaps across models. For each sample, we generate 128 tokens and repeat the measurement 3 times. The measured generation time covers vision processing and decoder generation, but excludes CPU preprocessing and input transfer to the GPU. Memory usage is reported as peak allocated GPU memory in GiB.

Table~\ref{tab:eval_all_models} extends the main-paper cost comparison to all 3 models under this protocol. Our method reduces generation time by approximately 32--33\% and memory usage by 57--59\% relative to BF16, showing consistent savings across models.

\paragraph{Evaluation note.}
During accuracy evaluation, we stop generation early when repetitive degeneration produces excessively long responses.

\section{Qualitative results}
To illustrate how quantization affects model responses, Figures~\ref{fig:qual_ai2d_chartqa}--\ref{fig:qual_nocaps_textcaps} present qualitative comparisons across the 8 evaluation tasks. Each example pairs the input with responses from the compared methods, allowing direct comparison of their answers under the indicated quantization settings.

\end{document}